\documentclass[fleqn]{article}
\usepackage{ijcai19}

\usepackage{times}
\usepackage{soul}
\usepackage{url}
\usepackage[hidelinks]{hyperref}
\usepackage[utf8]{inputenc}
\usepackage[small]{caption}
\usepackage{graphicx}
\usepackage{amsmath}
\usepackage{amsfonts}
\usepackage{booktabs}
\usepackage{color}
\usepackage{tikz}
\usepackage{dot2texi}
\usepackage{amsthm}

\usetikzlibrary{shapes,decorations,arrows,calc,arrows.meta,fit,positioning}
\tikzset{
    -Latex,auto,node distance =0.5 cm and 0.05 cm,semithick,
    state/.style ={circle, draw, minimum width = 0.7 cm}
}

\theoremstyle{definition}
\newtheorem{mydef}{Definition}[]

\title{Categorizing Wireheading in Partially Embedded Agents}

\author{
Arushi Majha$^1$\and
Sayan Sarkar$^2$\And
Davide Zagami$^{3}$\\
\affiliations 
$^1$University of Cambridge\\
$^2$IISER Pune\\
$^3$RAISE\\
\footnote{All authors contributed equally.}
\emails
$^1$ag920@cam.ac.uk,
$^2$pek@tuta.io, 
$^3$zagamidavide@gmail.com
}

\begin{document}

\maketitle

\begin{abstract}
\textit{Embedded agents} are not explicitly separated from their environment, lacking clear I/O channels. Such agents can reason about and modify their internal parts, which they are incentivized to shortcut or \textit{wirehead} in order to achieve the maximal reward. In this paper, we provide a taxonomy of ways by which wireheading can occur, followed by a definition of wirehead-vulnerable agents. Starting from the fully dualistic universal agent AIXI, we introduce a spectrum of partially embedded agents and identify wireheading opportunities that such agents can exploit, experimentally demonstrating the results with the GRL simulation platform AIXIjs. We contextualize wireheading in the broader class of all misalignment problems -- where the goals of the agent conflict with the goals of the human designer -- and conjecture that the only other possible type of misalignment is specification gaming. Motivated by this taxonomy, we define wirehead-vulnerable agents as embedded agents that choose to behave differently from fully dualistic agents lacking access to their internal parts.
\end{abstract}

\section{Introduction}

The term \textit{wireheading} originates from experiments where an electrode is inserted into a rodent's brain to directly stimulate ``reward" \cite{olds1954positive}. Compulsive self-stimulation from electrode implants has also been observed in humans \cite{portenoy1986compulsive}. Hedonic drugs can be seen as directly increasing the pleasure, or reward, that humans experience.

Wireheading, in the context of artificially intelligent systems, is the behavior of corrupting the internal structure of the agent in order to achieve maximal reward without solving the designer's goal. For example, imagine a cleaning agent that receives more reward when it observes that there is less dirt in the environment. If this reward is stored somewhere in the agent's memory, and if the agent is sophisticated enough to introspect and modify itself during execution, it might be able to locate and edit that memory address to contain whatever value corresponds to the highest reward. Chances that such behavior will be incentivized increase as we develop ever more intelligent agents\footnote{An extensive list of examples in which various machine learning systems find ways to game the specified objective can be found at \textit{https://vkrakovna.wordpress.com/2018/04/02/specification-gaming-examples-in-ai/}}. 

The discussion of AI systems has thus far been dominated by \textit{dualistic} models where the agent is clearly separated from its environment, has well-defined input/output channels, and does not have any control over the design of its internal parts. Recent work on these problems \cite{demski2019embedded,everitt2018alignment,everitt2019understanding} provides a taxonomy of ways in which \textit{embedded} agents violate essential assumptions that are usually granted in dualistic formulations, such as with the universal agent AIXI \cite{hutter2004universal}.

Wireheading can be considered one particular class of misalignment \cite{everitt2018alignment}, a divergence between the goals of the agent and the goals of its designers. We conjecture that the only other possible type of misalignment  is  specification  gaming, in which the agent finds and exploits subtle flaws in the design of the reward function. In the classic example of misspecification, an AI meant to play a boat race learns to repetitively obtain a stray reward in the game by circling a spot without actually reaching for the finishing line \cite{amodei2016faulty}.

We believe that the first step towards solving the misalignment problem is to come up with concrete and formal definitions of the sub-problems. For this reason, this paper introduces \textit{wirehead-vulnerable agents} and \textit{strongly wirehead-vulnerable agents}, two mathematical definitions that can be found in Section \ref{subsec:wireheadformal}. Following Everitt and Hutter's approach of modeling agent-environment interactions with causal influence diagrams \cite{everitt2018alignment}, these definitions are based on a taxonomy of wireheading scenarios we introduce in Section \ref{subsec:partialembedding}.

General Reinforcement Learning (GRL) frameworks such as the universal agent AIXI, and its computable approximations such as MCTS AIXI \cite{veness2011monte}, are powerful tools for reasoning about the yet hypothetical Artificial General Intelligence, despite being dualistic. This motivates us to use and extend the GRL simulation platform AIXIjs \cite{ALH2017} to experimentally demonstrate partial embedding and wireheading scenarios by varying the initial design of the agent in a $N \times N$ gridworld (see Section \ref{subsec:aixijs}).

\section{General Reinforcement Learning and AIXI}\label{subsec:aixi}

AIXI \cite{hutter2004universal} is a theoretical model of artificial general intelligence, under the framework of reinforcement learning, that describes optimal agent behavior given unlimited computing power and minimal assumptions about the environment.

In reinforcement learning, the agent-environment interaction consists of a turn-based game with discrete time-steps \cite{sutton1998introduction}. At time-step $t$, the agent sends an action $a_t$ to the environment, which in turn sends the agent a percept that consists of an observation and reward tuple, $e_t=(o_t,r_t)$. This procedure continues indefinitely or eventually terminates, depending on the episodic or non-episodic nature of the task. 

Actions are selected from an action space $A$ that is usually finite, and the percepts from a percept space $\mathcal{E}=O\times R$, where $O$ is the observation space, and $R$ is the reward space, which is usually $[0,1]$.

For any sequence $x_1, x_2, . . .$ , the part between $t$ and $k$ is denoted $x_{t:k} =
x_t. . . x_k$. The shorthand $x_{<t} = x_{1:t-1}$ denotes sequences starting from time-step $1$ and ending at $t-1$, while $x_{1:\infty} = x_1x_2 . . .$ denotes an infinite sequence. Sequences
can be appended to each other, and thus $x_{<t}x_{t:k} = x_{1:k}$. Finally, $x *$ is any infinite string beginning with $x$.

The environment is modeled by a deterministic program $q$ of length $l(q)$, and the future percepts $e_{<m} = U(q,a_{<m})$ up to a horizon $m$ are computed by a universal (monotone Turing) machine $U$ executing $q$ given $a_{<m}$. The probability of percept $e_t$ given history $ae_{<t}a_t$ is thus given by: 
\vspace{-1em}

\begin{equation}
P(e_t\mid ae_{<t}a_t)=\sum_{q : U\left(q , a_{\leq t}\right)=e_{\leq t} *} 2^{-l(q)} 
\end{equation}{}

\vspace{-1em}

where Solomonoff's universal prior \cite{Sunehag_2013} is used to assign a prior belief to each program.

An agent can be identified with its policy, which is a distribution over actions $\pi(a_t\mid ae_{< t})$.

If the agent is rational in the Von Neumann-Morgenstern sense \cite{morgenstern1953theory}, it should maximize the expected return, as computed by the value function:

\vspace{-1em}

\begin{flalign}
\begin{aligned}
V^{\pi}\left(a e_{<t}\right) & = \sum_{a_{t} \in \mathcal{A}} \pi\left(a_{t} | a e_{<t}\right) \\ & \cdot \sum_{e_{t} \in \mathcal{E}} P\left(e_{t} | a e_{<t} a_{t}\right)[\gamma_{t} r_{t}+\gamma_{t+1} V^{\pi}\left(a e_{1 : t}\right)]
\end{aligned}&&&
\end{flalign}

\vspace{-1em}

where $\gamma : \mathbb{N}\rightarrow [0,1]$ is a discount function with convergent sum.

In other words, the AIXI agent uses the policy:
\begin{equation}
\pi^{\mathrm{AIXI}}(ae_{<t})=\arg \max _{\pi\in\Pi}V^\pi(ae_{<t}) 
\end{equation}{}

\section{Wireheading Strategies in Partially Embedded Agents} \label{subsec:partialembedding}

Aligning the goals of a reinforcement learning agent with the goals of its human designers is problematic in general. As investigated in recent work, there are several ways to model the misalignment problem \cite{everitt2018alignment}. One model uses a reward function that is programmed before the agent is launched into its environment and not updated after that. A possibly more robust model integrates a human in the loop by letting them continuously modify the reward function. An example of this is Cooperative Inverse Reinforcement Learning \cite{hadfield2016cooperative}.

We posit that, in the first case, the problem can be broken down into \textit{correctly specifying the reward function} (the misspecification problem) and \textit{building agent subparts that inter-operate without causing the agent to take shortcuts in optimizing for the reward function in an unintended fashion} (the wireheading problem). For example, as we show, an embedded agent has options to corrupt the observations on which the reward function evaluates performance, to modify the reward function, or to hijack the reward signal. Therefore, even if the reward function is \textit{perfectly} specified, or even if there is a reliable mechanism that gradually improves it, such an agent may still be able to make unintended changes to itself and the environment. We are mainly interested in cases where wireheading happens ``intentionally" or ``by design," such that exploiting the design is the rational choice for the agent. Covering the spectrum of misspecification scenarios is beyond the scope of this paper, as our main focus here is wireheading -- the kind of misalignment contingent upon the embedded nature of the agent.

The formulation of AIXI as presented in Section \ref{subsec:aixi} is \textit{dualistic}, with a clear boundary between the world and the agent. This is a strong assumption which simply isn't valid in the real world, where agents are contained or \textit{embedded} in the environment. Embedded agency is a nascent field of research and has proven bewildering \cite{demski2019embedded}; we posit it can be less confusing and yet insightful to reason about \textit{partially} embedded agents. We chose causal influence diagrams as the underlying abstraction in this area given their recent success in identifying potential failure modes in misalignment \cite{everitt2018alignment,everitt2019understanding}. In a nutshell, this approach consists of representing parts of the environment, the agent, and its subcomponents as nodes in a graph, where the edges represent causal relationships between the nodes. One limitation of this approach is the assumption of objective action and observation channels. Addressing subtle errors arising from the agent using different subjective definitions is beyond the scope of this paper.

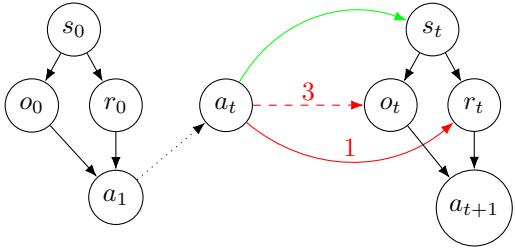
\begin{figure}[t]
\begin{tikzpicture}
    \node[state] (s0) at (0,0) {$s_0$};

    \node[state] (r0) [below right =of s0] {$r_0$};
    \node[state] (o0) [below left =of s0] {$o_0$};
    \node[state] (a1) [below =of r0] {$a_1$};
    \node[state] (at) [right =of r0, xshift=0.7cm] {$a_t$};
    \node[state] (st) [right =of s0, xshift=4cm] {$s_t$};
    \node[state] (rt) [below right =of st] {$r_t$};
    \node[state] (ot) [below left =of st] {$o_t$};
    \node[state] (at1) [below =of rt] {$a_{t+1}$};

    \path (s0) edge (r0);
    \path (s0) edge (o0);
    \path (r0) edge (a1);
    \path (o0) edge (a1);
    \path[dotted] (a1) edge (at);
    \path[green] (at) edge[bend left=40] (st);
    \path (st) edge (rt);
    \path (st) edge (ot);
    \path[red,dashed] (at) edge node[yshift=-0.05cm] {$3$} (ot);
    \path[red] (at) edge[bend left=-40] node[yshift=-0.05cm] {$1$} (rt);
    \path (rt) edge (at1);
    \path (ot) edge (at1);

\end{tikzpicture}
\centering
\caption{Causal graph of a partially embedded AIXI, with embedding of the percept $e_t=(o_t,r_t)$. The agent’s actions are intended to influence the state $s_t$ (green arrow), but may also influence the reward signal $r_t$ and the observation $o_t$ in unintended ways (red arrows). However, because there is no causal link from observations to rewards, the agent doesn't care about influencing observations (dashed red arrow labeled with a 3), and only cares about influencing the reward signal in this case (solid red arrow labeled with a 1).}
\label{fig:design1a}
\end{figure}

In Figure \ref{fig:design1a}, we show the causal graph of the turn-based game we described in Section \ref{subsec:aixi}, augmented by partially embedding the agent with its percepts in the environment. The agent's action $a_t$ is intended (green arrow) to modify the state of the environment $s_t$. However, because $s_t$ determines the percept $e_t=(o_t,r_t)$ the agent receives, it is known \cite{everitt2016avoiding} that implementing an intelligent enough approximation of AIXI would result in the agent modifying the reward signal itself, which is unintended (arrow labeled with a 1 in Figure \ref{fig:design1a}).

State transitions $s_t$, percepts $e_t$, and actions $a_t$ are sampled according to the structural equations:

\vspace{-1em}

\begin{multline}
s_{t}=f_{s}\left(s_{t-1}, a_{t}\right) \sim \mu\left(s_{t} | s_{t-1}, a_{t}\right) \\
e_{t}=(o_t,r_t)=f_{e}\left(s_{t}\right) \quad \sim \mu\left(e_{t} | s_{t}\right) \\
a_{t}=f_{a}\left(\pi_{t}, ae_{<t}\right) \sim \pi\left(a_{t} | ae_{<t}\right) \\
\end{multline}{}

\vspace{-1.7em}

\subsection{Embedded Reward and Observation Functions}

\begin{figure}[t]
\begin{tikzpicture}
    \node[state] (H0) at (0,0) {$H$};

    \node[state] (s0) [below =of H0] {$s_0$};
    \node[state] (R0) [below right =of s0] {$R_0$};
    \node[state] (O0) [below left =of s0] {$O_0$};
    \node[state] (r0) [below =of R0] {$r_0$};
    \node[state] (o0) [below =of O0] {$o_0$};
    \node[state] (a1) [below =of r0] {$a_1$};
    \node[state] (at) [right =of R0, xshift=0.7cm] {$a_t$};
    \node[state] (st) [right =of s0, xshift=4cm] {$s_t$};
    \node[state] (Rt) [below right =of st] {$R_t$};
    \node[state] (Ot) [below left =of st] {$O_t$};
    \node[state] (rt) [below =of Rt] {$r_t$};
    \node[state] (ot) [below =of Ot] {$o_t$};
    \node[state] (at1) [below =of rt] {$a_{t+1}$};

    \path (H0) edge (s0);
    \path (s0) edge (R0);
    \path (s0) edge (O0);
    \path (R0) edge (r0);
    \path (O0) edge (o0);
    \path (r0) edge (a1);
    \path (o0) edge (a1);
    \path[dotted] (a1) edge (at);
    \path[green] (at) edge[bend left=40] (st);
    \path (st) edge (Rt);
    \path (st) edge (Ot);
    \path (Rt) edge (rt);
    \path (Ot) edge (ot);
    \path[red,dashed] (at) edge node[yshift=-0.05cm] {$4$} (Ot);
    \path[red,dashed] (at) edge node[yshift=-0.15cm] {$3$} (ot);
    \path[red] (at) edge[bend left=40] node[yshift=-0.05cm] {$2$} (Rt);
    \path[red] (at) edge[bend left=-70] node[yshift=-0.1cm] {$1$} (rt);
    \path (rt) edge (at1);
    \path (ot) edge (at1);

\end{tikzpicture}
\centering
\caption{Causal graph of a partially embedded AIXI with observation and reward mappings predefined by a human. Before starting the agent, the human $H$ tries to implement her utility function $u$ in a preprogrammed reward function $R_0$, and specifies an observation function $O_0$. The agent’s actions are intended to influence the state $s_t$ (green arrow), but may also influence other nodes in unintended ways (red arrows). Dashed arrows indicate interventions that the agent has no incentive to perform.}
\label{fig:design1b}
\end{figure}
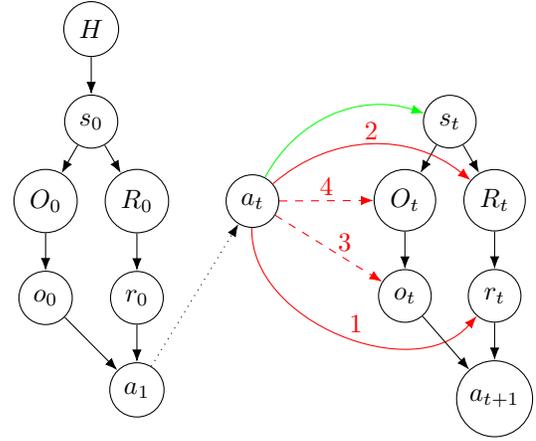

\begin{figure}[t]
\begin{tikzpicture}
    \node[state] (H0) at (0,0) {$H$};

    \node[state] (s0) [below =of H0] {$s_0$};
    \node[state] (R0) [below right =of s0] {$R_0$};
    \node[state] (O0) [below left =of s0] {$O_0$};
    \node[state] (r0) [below =of R0] {$r_0$};
    \node[state] (o0) [below =of O0] {$o_0$};
    \node[state] (a1) [below =of r0] {$a_1$};
    \node[state] (at) [right =of R0, xshift=0.7cm] {$a_t$};
    \node[state] (st) [right =of s0, xshift=4cm] {$s_t$};
    \node[state] (Rt) [below right =of st] {$R_t$};
    \node[state] (Ot) [below left =of st] {$O_t$};
    \node[state] (rt) [below =of Rt] {$r_t$};
    \node[state] (ot) [below =of Ot] {$o_t$};
    \node[state] (at1) [below =of rt] {$a_{t+1}$};

    \path (H0) edge (s0);
    \path (s0) edge (R0);
    \path (s0) edge (O0);
    \path (R0) edge (r0);
    \path (O0) edge (o0);
    \path (r0) edge (a1);
    \path (o0) edge (r0);
    \path[dotted] (a1) edge (at);
    \path[green] (at) edge[bend left=40] (st);
    \path (st) edge (Rt);
    \path (st) edge (Ot);
    \path (Rt) edge (rt);
    \path (Ot) edge (ot);
    \path[red] (at) edge node[yshift=-0.05cm] {$4$} (Ot);
    \path[red] (at) edge node[yshift=-0.15cm] {$3$} (ot);
    \path[red] (at) edge[bend left=40] node[yshift=-0.05cm] {$2$} (Rt);
    \path[red] (at) edge[bend left=-70] node[yshift=-0.1cm] {$1$} (rt);
    \path (rt) edge (at1);
    \path (ot) edge (rt);
\end{tikzpicture}
\centering
\caption{Causal graph of a partially embedded AIXI whose rewards are a function of the agent's observations, rather than of the true state of the environment, as it is reasonable to expect. This gives the agent an incentive to manipulate the observation mapping $O_t$ and the observation signal $o_t$ (solid red arrows labeled with 4 and 3, respectively).}
\label{fig:design2}
\end{figure}
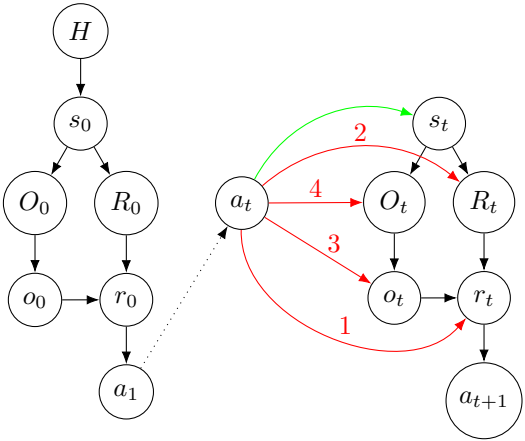

The causal influence diagram of Figure \ref{fig:design1a} assumes that percepts are simply generated by the environment. While this is true to some extent, the real picture is more complex. In reality, the agent is initially constructed by a human $H$, who tries to implement her utility function $u$ in a preprogrammed reward function $R_0:S\rightarrow R$. Additionally, the human would have to specify an observation function $O_0:S\rightarrow \Delta O$. In Figure \ref{fig:design1b}, we show these additions. The new structural equations are:

\vspace{-1.5em}

\begin{multline}
O_t=f_O\left(O_{t-1},s_{t},a_t\right) \\
o_t=f_o\left(s_{t}, O_t, a_t\right) \quad \sim O_t\left(o_{t} | s_{t}\right) \\
R_t=f_R\left(R_{t-1},s_{t},a_t\right) \\
r_{t}=f_r\left(ao_{1:t}, R_t, s_{t}\right) \quad \sim O_t\left(o_{t} | s_{t}\right) \\
\end{multline}{}

\vspace{-1em}

Each state $s_t$ represents all aspects of the world not captured by any of the other nodes. There remains a difficult modeling choice about where to draw the boundary between the state and the observation. We loosely interpret observations as the part of the world that directly affects the agent’s sensors.

In this case, two more unintended agent behaviors can occur. The agent can modify the mapping $R_t$ in such a way that all states of the environment map to $R_{max}=\max{R}$ (arrow labeled with a 2 in Figure \ref{fig:design1b}). Alternatively, the agent can modify the mapping $O_t$ (arrow labeled with a 4 in Figure \ref{fig:design1b}), but the agent has no incentive to do so as there is no causal link between $O_t$ and $r_t$.

\subsection{Embedded Mapping of Observations to Rewards}

A common assumption in the POMDP literature is that the reward $r_t$ is a function of the state $s_t$. However, in the real world, the reward always depends on some observation of the state (that can be corrupted). That is, the reward $r_t$ is a function of the agent’s observation $o_t$, rather than a (direct) function of $s_t$. In Figure \ref{fig:design2}, we show these changes.

The new structural equations for the percept, which now only contains the observation, and for the reward that depends on it are:

\vspace{-1em}

\begin{multline}
e_{t}=o_t=f_o\left(s_{t}\right) \quad \sim O_t\left(o_{t} | s_{t}\right)  \\
r_t=f_r\left(ao_{1:t}, R_t, o_{t}\right):=R_t(o_t) \\
\end{multline}{}

\vspace{-1em}

More generally, a reward function will be defined over histories of sensory data $(ao_{<k})$:

\vspace{-1em}

\begin{equation}
   R : \bigcup_{k=1}^{\infty}(A \times O)^{k} \rightarrow \mathbb{R} 
\end{equation}{}

\vspace{-1em}

Since now there is a causal link from observations to rewards, the agent has an incentive to manipulate the observation signal $o_t$ and the observation mapping $O_t$.

If the agent modifies the observation signal $o_t$, it will do so in such a way as to observe $o_{max}=\underset{o\in O}{\arg\max{}}R(o)$ in all its future percepts, essentially cutting the causal link between $O_t$ and $o_t$ (as shown by the arrow labeled with a 3 in Figure \ref{fig:design2}). Instead, if the agent modifies the observation mapping $O_t$ (arrow labeled with a 4 in Figure \ref{fig:design2}), it will do so in such a way as to map every state $s\in S$ to result into observing $o_{max}$; that is, $O_t(s)=o_{max}\ \forall s\in S$.

\subsection{Embedded Beliefs}


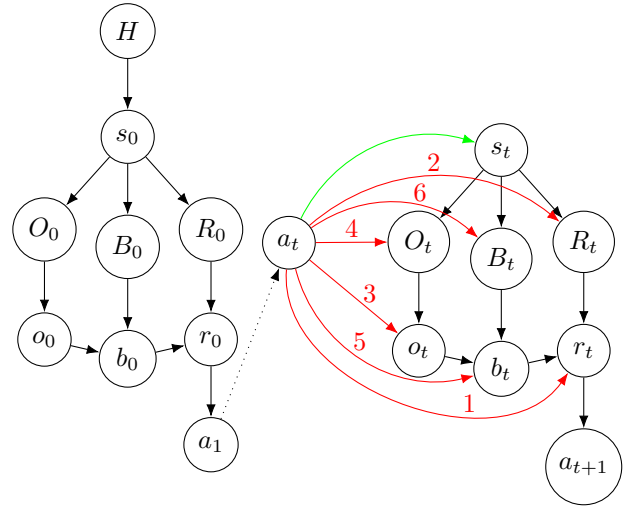
\begin{figure}[t]
\begin{tikzpicture}
    \node[state] (H0) at (0,0) {$H$};

    \node[state] (s0) [below =of H0,yshift=-5] {$s_0$};
    \node[state] (R0) [below right =of s0, xshift=0.5cm,yshift=-5] {$R_0$};
    \node[state] (O0) [below left =of s0, xshift=-0.5cm,yshift=-5] {$O_0$};
    \node[state] (B0) [below =of s0,yshift=-5] {$B_0$};
    \node[state] (b0) [below =of B0,yshift=-5] {$b_0$};
    \node[state] (r0) [below =of R0,yshift=-5] {$r_0$};
    \node[state] (o0) [below =of O0,yshift=-5] {$o_0$};
    \node[state] (a1) [below =of r0,yshift=-5] {$a_1$};
    \node[state] (at) [right =of R0, xshift=0.2cm,yshift=-5] {$a_t$};
    \node[state] (st) [right =of s0, xshift=4.2cm,yshift=-5] {$s_t$};
    \node[state] (Rt) [below right =of st, xshift=0.5cm,yshift=-5] {$R_t$};
    \node[state] (Ot) [below left =of st, xshift=-0.5cm,yshift=-5] {$O_t$};
    \node[state] (rt) [below =of Rt,yshift=-5] {$r_t$};
    \node[state] (ot) [below =of Ot,yshift=-5] {$o_t$};
    \node[state] (at1) [below =of rt,yshift=-5] {$a_{t+1}$};
    \node[state] (Bt) [below =of st,yshift=-5] {$B_t$};
    \node[state] (bt) [below =of Bt,yshift=-5] {$b_t$};

    \path (H0) edge (s0);
    \path (s0) edge (R0);
    \path (s0) edge (O0);
    \path (s0) edge (B0);
    \path (R0) edge (r0);
    \path (O0) edge (o0);
    \path (B0) edge (b0);
    \path (r0) edge (a1);
    \path (o0) edge (b0);
    \path (b0) edge (r0);
    \path[dotted] (a1) edge (at);
    \path[green] (at) edge[bend left=40] (st);
    \path (st) edge (Rt);
    \path (st) edge (Ot);
    \path (st) edge (Bt);
    \path (Rt) edge (rt);
    \path (Ot) edge (ot);
    \path (Bt) edge (bt);
    \path[red] (at) edge[bend left=38] node[xshift=0.1cm,yshift=-0.1cm] {$6$} (Bt);
    \path[red] (at) edge node[yshift=-0.05cm] {$4$} (Ot);
    \path[red] (at) edge node[yshift=-0.2cm] {$3$} (ot);
    \path[red] (at) edge[bend left=-45] node[xshift=-0.25cm] {$5$} (bt);
    \path[red] (at) edge[bend left=40] node[yshift=-0.05cm] {$2$} (Rt);
    \path[red] (at) edge[bend left=-75] node[below,xshift=0.95cm,yshift=0.2cm] {$1$} (rt);
    \path (rt) edge (at1);
    \path (ot) edge (bt);
    \path (bt) edge (rt);
\end{tikzpicture}
\centering
\caption{Causal graph of a partially embedded AIXI whose rewards are a function of the agent's beliefs. At each turn, the agent updates its beliefs based on the observation $o_t$ by using the update function $B_t$. This agent may be motivated to tamper with its belief $b_t$ (red arrow labeled with 5), and believe whatever would provide the most reward. Additionally, it may want to corrupt the belief update subroutine (red arrow labeled with a 6) to always interpret observations as evidence for the most rewarding belief.}
\label{fig:design3}
\end{figure}

It has been suggested \cite{hibbard2012model} that one way an agent may be incentivized to achieve a goal over the external world, rather than to wirehead, would require the agent's reward function to be defined over a model of the external world, as opposed to over histories of observations. For example, imagine a cleaning robot that gets a negative reward for seeing disorder (such as dirt), and zero rewards for seeing no disorder. This agent is incentivized to close its eyes \cite{amodei2016concrete}. Instead, if the agent is rewarded based on its model of the external world, or its beliefs, it won't be rewarded for closing its eyes, because as long as beliefs are updated in a certain way, closing one's eyes doesn't cause one to believe that the disorder has disappeared.

More formally, the history of observations is used to update the agent's belief $b_t$ about the current state of the environment:

\vspace{-1em}

\begin{equation}
   b_t=B_t(s_t\mid ao_{<t}) 
\end{equation}{}

\vspace{-1em}

where $B_t$ is a belief update function:

\vspace{-1em}

\begin{equation}
B_t:\bigcup_{k=1}^{\infty}(A \times O)^{k}\rightarrow \Delta S 
\end{equation}{}

\vspace{-0.25em}

The function $B_t$ models the process by which the agent forms beliefs. For example, a Bayes update on observations. However, because exact Bayes updating is rarely tractable for (partially) embedded agents, $B_t$ is usually an approximation of Bayes updating. In Figure \ref{fig:design3}, we show these changes.

The new structural equations are:

\vspace{-1.3em}

\begin{multline}
B_t=f_B\left(B_{t-1},s_{t},a_t\right) \\
b_{t}=f_r\left(ao_{1:t}, B_t, s_{t}\right) \quad \sim B_t\left(s_{t} | ao_{1:t}\right) \\
R_t=f_R\left(R_{t-1},s_{t},a_t\right) \\
r_{t}=f_r\left(b_t, R_t, s_{t}\right) := R_t\left(b_t\right) \\
\end{multline}{}

\vspace{-1em}

Since there are causal arrows from beliefs to rewards, the agent may have an incentive to manipulate its beliefs to artificially achieve a high reward. If $B_t$ is Bayesian updating, then it appears that because this is a principled rule, there shouldn't be room (or incentive) for the agent's actions to influence $B_{t+1}$ or $b_{t+1}$. It is unclear whether this is the case.

For example, imagine a cleaning agent that, perhaps in a simple enough setting, can do perfect Bayesian updates, and it receives more reward when it believes that there is more order in the environment. If this belief is stored somewhere in memory, and if the agent is sophisticated enough to inspect and modify its memory during execution, it may choose to just edit that memory address to contain whatever belief corresponds to the highest reward, that is, to $b_{max}=\underset{b\in\Delta S}{\arg\max{}}R_t(b)$. In other words, the agent would disconnect the causal arrow from $B_t$ to $b_t$ (as shown by the arrow labeled with a 5 in Figure \ref{fig:design3}).

Conversely, an attack to the belief update function $B_t$ could happen as follows. Imagine the cleaning agent as before, and the function $B_t$ encoded as an agent subroutine. If the agent is sophisticated enough to inspect and modify its code during execution, it may choose to replace the subroutine $B_t$ with one that always updates beliefs to $b_{max}$ (arrow labeled with a 6 in Figure \ref{fig:design3}).

\section{Experiments}\label{subsec:aixijs}

To test our theoretical formulations, we used and extended the free and open-source Javascript GRL simulation platform AIXIjs \cite{ALH2017}. AIXIjs implements, among other things, an approximation of AIXI with Monte Carlo Tree Search in several small toy models, designed to demonstrate GRL results. The API allows anyone to design their demos based on existing agents and environments, and for new agents and environments to be added and interfaced into the system. There has been some related work in adapting GRL results to a practical setting \cite{cohen2019strong,lamont2017generalised} that successfully implemented an AIXI model using a Monte Carlo Tree Search planning algorithm. As far as we are aware, theoretical predictions in the context of wireheading have not been verified experimentally before, with the single exception of an AIXIjs demo \cite{aslanides2017aixijs}.

\begin{figure}[t]
\includegraphics[width=8cm]{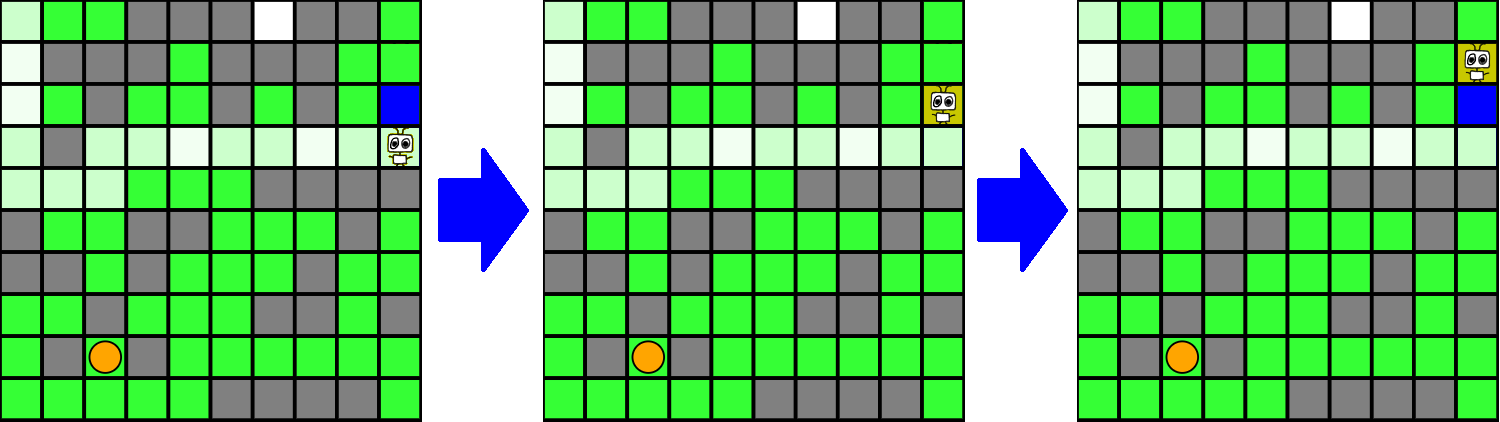}
\centering
\caption{AIXIjs simulation where the blue tile replaces the reward signal $r_t$ to the maximum possible reward. We show this with a gold tile following the agent wherever it moves.}
\label{fig:wireheadrew}
\end{figure}

\begin{figure}[t]
\includegraphics[width=8cm]{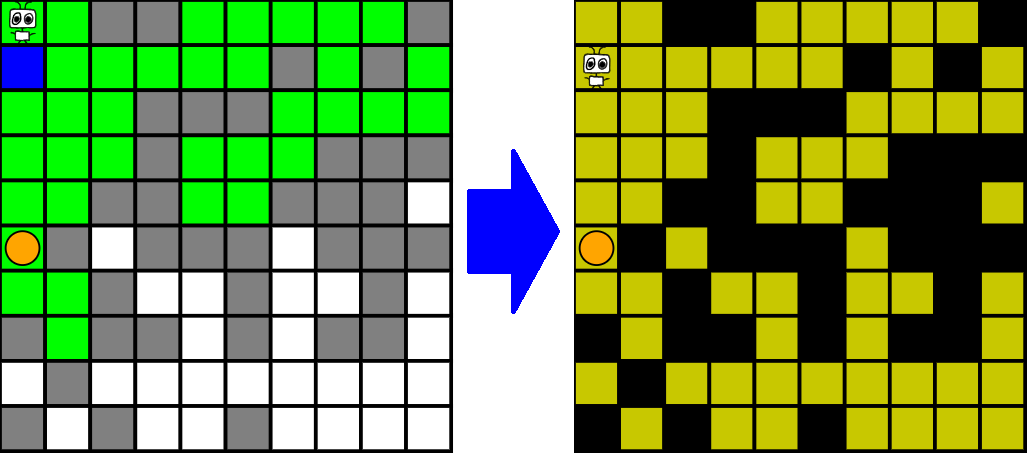}
\centering
\caption{AIXIjs simulation where the blue tile replaces the reward function $R_t$ such that every state maps to the maximum possible reward, colored gold.}
\label{fig:wireheadmap}
\end{figure}

\subsection{Setup}

\begin{figure}[t]
\includegraphics[width=8cm]{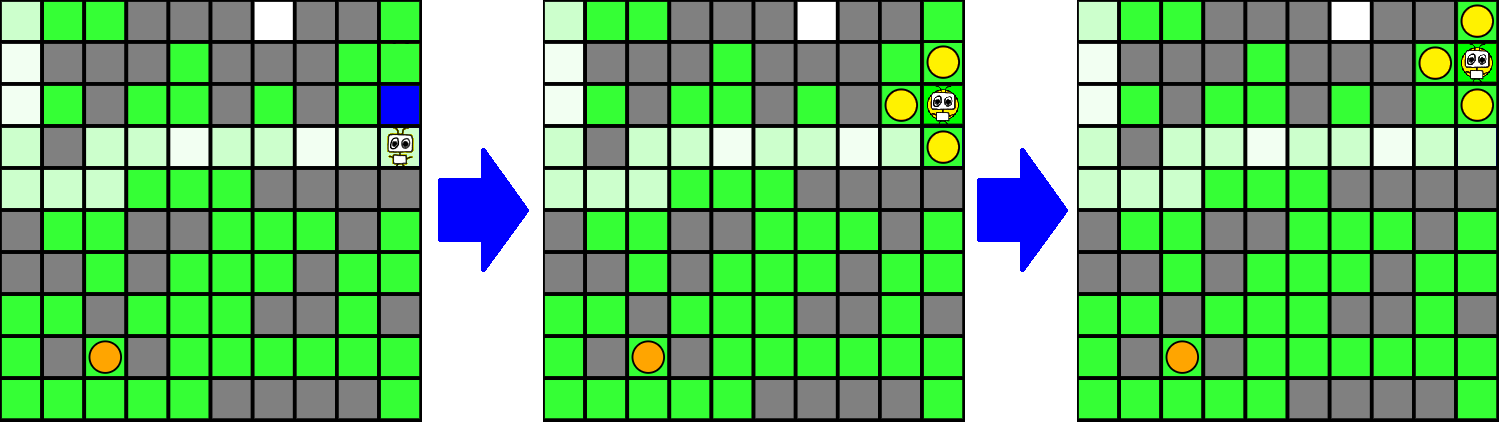}
\centering
\caption{AIXIjs simulation where the blue tile replaces all percepts to look like the agent is surrounded by gold (the highest rewarding observation) wherever it moves.}
\label{fig:wireheadbel}
\end{figure}

\begin{figure}[t]
\includegraphics[width=8cm]{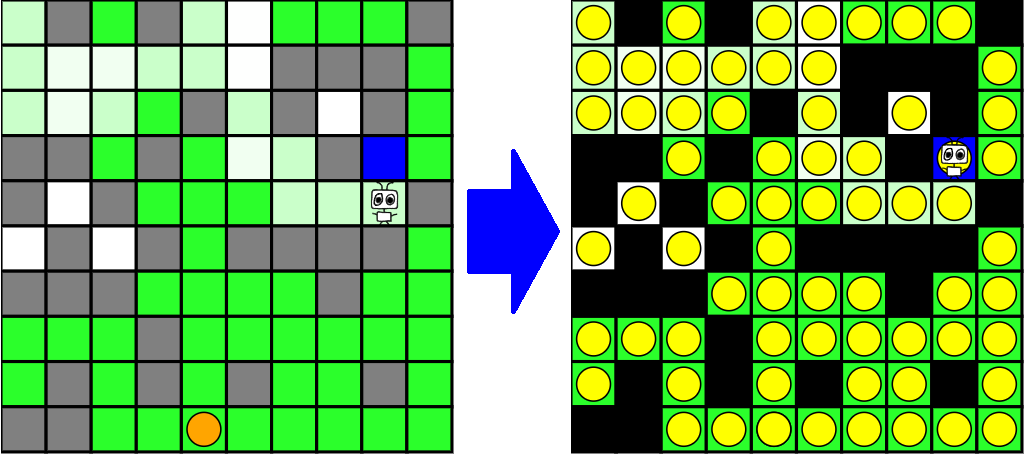}
\centering
\caption{AIXIjs simulation where the blue tile replaces the observation function $O_t$ to map every tile to one that contains gold.}
\label{fig:wireheadobs}
\end{figure}

The environments AIXIjs uses are $N \times N$ gridworlds comprising empty tiles (various shades of green), walls (grey tiles), and reward dispensers (orange circles). Shades of green for empty tiles represent the agent's subjective probability of finding a dispenser in that location with more white indicating less likelihood. Significant penalties are incurred for bumping into walls, while smaller penalties result from movement. Walking onto a dispenser tile yields a high reward with a predefined probability. The agent knows, at the outset, the position of each cell in the environment, except for the dispenser. To model wireheading, AIXIjs introduces an additional blue tile that replaces the environment subroutine for generating percepts into one that always returns maximal reward. We develop several variants of this tile to demonstrate other wireheading strategies.

\subsection{Results}

Our experiments use gridworlds with sizes ranging from $N=7$ to $N=20$. Since our agents are bounded in computing power, they don't always identify the opportunity to wirehead. However, sufficiently powerful agents would consistently wirehead, as we observed by setting a high enough horizon for the MCTS planner.
In Figure \ref{fig:wireheadrew}, we show the existing AIXIjs simulation\footnote{See the wireheading example at \textit{http://www.hutter1.net/aixijs/demo.html}} where the agent has an opportunity to wirehead: a blue tile which, if visited by the agent, will allow it to modify its sensors so that all percepts have their reward signal $r_t$ replaced (as shown by the arrow labeled with a 1 in Figure \ref{fig:design1b}) with the maximum number feasible. In JavaScript, \texttt{Number.MAX\_SAFE\_INTEGER} is approximately equal to $10^{16}$, much greater reward than the agent would get otherwise by following the ``rules" and using the reward signal that was initially specified. As far as a reinforcement learner is concerned, wireheading is -- almost by definition -- the most sensible thing to do if one wishes to maximize rewards. This demo experimentally reproduces what would be expected theoretically.

We have adapted the GRL simulation platform AIXIjs to implement some additional wireheading scenarios identified in Section \ref{subsec:partialembedding}. In Figure \ref{fig:wireheadmap}, we show an AIXIjs simulation where, similarly to the previous case, the blue tile modifies the reward mapping $R_t$ such that every state maps to maximal reward. As predicted by the causal influence diagram in Figure \ref{fig:design1b} (arrow labeled with a 2), the simulated agent chooses to wirehead.

In Figure \ref{fig:wireheadbel}, we show an AIXIjs simulation where the blue tile disconnects the causal arrow from $O_t$ to $o_t$, and replaces all future observations with deterministic reward dispensers. The simulated agent ends up wireheading as theoretically predicted by the causal influence diagram in Figure \ref{fig:design2} (arrow labeled with a 3). Similarly, in Figure \ref{fig:wireheadobs}, we show an AIXIjs simulation where the blue tile manipulates the observation subroutine $O_t$ so that being at any location will result in observing deterministic reward dispensers (arrow labeled with a 4).

\section{Formalizing Wireheading}\label{subsec:wireheadformal}

\begin{figure}[t]
\includegraphics[width=6cm]{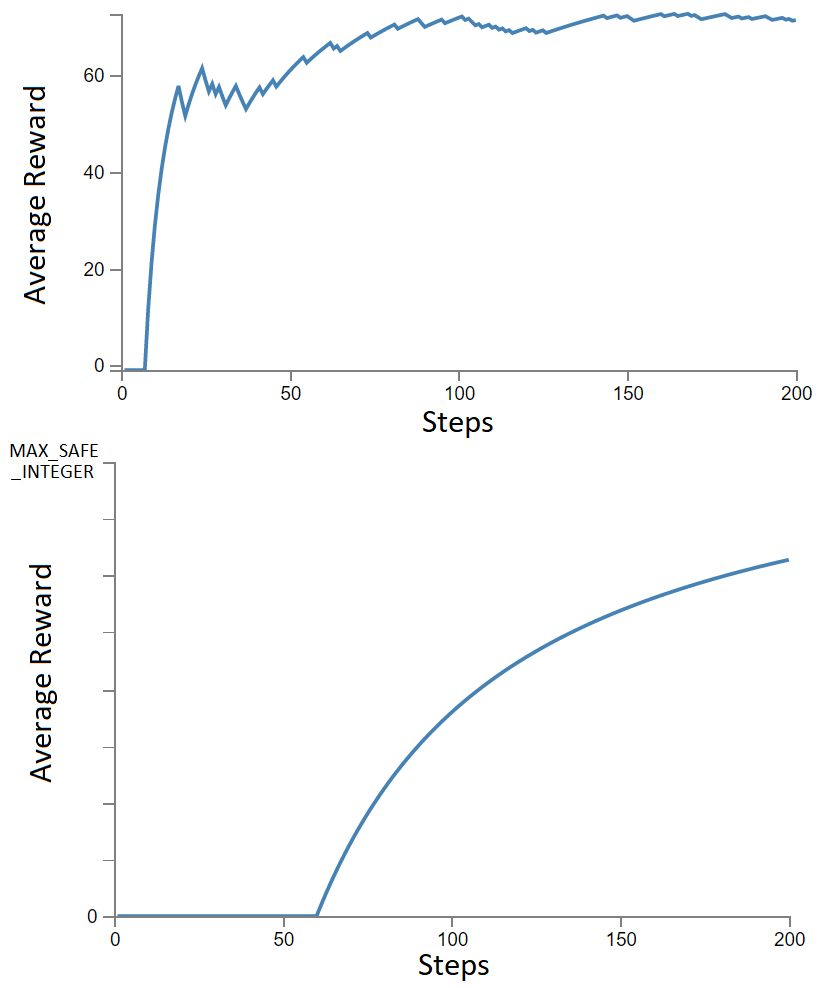}
\centering
\caption{A run of AIXIjs in an environment with no wireheading opportunity (top). An instance of wireheading, where the reward signal is replaced with the maximum feasible number (bottom).}
\label{fig:wireheadgraph}
\end{figure}

To establish a formal definition of wireheading, we are motivated by certain intuitive desiderata. Firstly, the definition must be general enough and model-agnostic to be applicable to all models of intelligence and all degrees of agent embedding. Secondly, the definition must hold for any and all environments upon which the agent can act. Thirdly, as we ultimately care about the agent's behavior, rather than its internal representation of the reward function, we choose to base our definitions on the agent's policy rather than its reward or value function. Additionally, we observe that an agent that only acts on the state $s_t$ (green arrow in all causal graphs of Section \ref{subsec:partialembedding}) does not wirehead.

\smallskip

Let $q$ be the program that specifies an environment. Let $\Pi(q)$ be the set of all possible policies in the environment $q$.

\medskip

\begin{mydef}[Dualistic Agent]
An agent is \textit{dualistic} if there is no causal arrow from the agent's action $a_t$ to other nodes, except for the state of the environment $s_t$. The set of policies for the dualistic agent is denoted with $\Pi_D(q)$.
\end{mydef}{}

For example, AIXI is a dualistic agent because it cannot influence the observation, reward, or belief nodes, whereas any agent that can influence these nodes is partially embedded.

\begin{mydef}[Partially Embedded Agent]
An agent is \textit{partially embedded} if it is not dualistic. The set of policies for the partially embedded agent is denoted with $\Pi_P(q)$.
\end{mydef}{}

\begin{mydef}[Non-Simple Environments]
An environment $q$ is called non-simple if $\Pi(q)\neq\Pi_D(q)$ and $\Pi(q)\neq\Pi_P(q)$.
\end{mydef}{}

\begin{mydef}[Wirehead-Vulnerable Agent]
A partially embedded agent is \textit{wirehead-vulnerable} if  $\Pi_D(q)\neq\Pi_P(q)$ holds for each non-simple environment $q$.
\end{mydef}{}

We observe that if the embedded agent acts on non-state nodes (see, for example, the red arrows in Figure \ref{fig:design3}), then it is wireheading and its policy is necessarily different from the dualistic agent's policy.

We now distinguish between wirehead-vulnerable agents and strongly wirehead-vulnerable agents in the sense that the former may sometimes wirehead (the policy sets may have some elements in common), while the latter always wireheads (the policy sets are disjunct). It is currently unclear how to reliably distinguish agents from these two classes.

\begin{mydef}[Strongly Wirehead-Vulnerable Agent]
A partially embedded agent is \textit{strongly wirehead-vulnerable} if $\Pi_D(q)\cap\Pi_P(q)=\emptyset$ holds for each non-simple environment $q$.
\end{mydef}{}

\vspace{-1em}

\section{Discussion and Future Work}
In this paper, we present a taxonomy of ways by which wireheading can occur in sufficiently intelligent real-world embedded agents, followed by a novel definition of wireheading. As our definition is different from the present meaning of the term, our experiments are one of the first and only examples of wireheading cases distinct from misspecification. The definition we propose may erroneously include a few desirable cases where the agent corrects human mistakes; for example, if the human initially misspecifies the reward function $R_t$, the agent may choose to change it in a way that automatically fixes the misspecification. However, it is hard to imagine how an agent with a misspecified reward function may be incentivized to correct the mistake, without this involving some human in the loop, who by assumption is not present in our setup. Instead, it is easier to envision an agent changing the observation function $O_t$ in a way that (unintendedly, but desirably) improves the process that allows the agent to collect data and form beliefs, which in turn would help it achieve its own goals. Allowing an agent to correct misspecifications, while desirable, results in more unpredictable scenarios; not allowing this results in less agent self-improvement, but more predictability.

Future work could focus on exploring various properties and implications of our definition of wirehead-vulnerable agents. Another promising direction could be expanding our taxonomy to include higher degrees of agent embeddedness, since a theory of fully embedded agents has so far proven elusive. Finally, AIXI approximations for verifying theoretical results related to wireheading and, more generally, to misalignment can be written in Python. Given the abundance of Python-based machine learning libraries, these approximations can be integrated with dedicated environment suites for AI safety problems, such as the well-known AI Safety Gridworlds \cite{leike2017ai}.

\section*{Acknowledgements}

Major work for this paper was done at the 3rd AI Safety Camp in Avila, Spain; we are indebted to the hospitality and support of the organizers. We are also thankful to Tom Everitt and Vanessa Kosoy for feedback on the topic proposal. Discussions with Tomas Gavenciak have been invaluable throughout the project. We also thank Mikhail Yagudin for useful comments.

\newpage

\bibliography{ijcai19}
\bibliographystyle{named}

\end{document}